# Enhanced Mortality Prediction In Patients With Subarachnoid Haemorrhage Using A Deep Learning Model Based On The Initial CT Scan

Sergio García-García MD,PhD[1*]; Santiago Cepeda MD,PhD[1*]; Dominik Müller PhD[2]; Alejandra Mosteiro MD[3]; Ramón Torné MD,PhD[3]; Silvia Agudo BSc[1]; Natalia de la Torre BSc[1]; Ignacio Arrese MD,PhD[1]; Rosario Sarabia MD,PhD[1].

1.- Neurosurgery Department, Rio Hortega University Hospital, Valladolid, Spain.

2.- IT-Infrastructure for Translational Medical Research, University of Augsburg, Augsburg, Germany.

3.- Neurosurgery Department, Hospital of Barcelona, Barcelona, Spain.

* These authors contributed equally

# ABSTRACT

PURPOSE: Subarachnoid hemorrhage (SAH) entails high morbidity and mortality rates. Convolutional neural networks (CNN), a form of deep learning, are capable of generating highly accurate predictions from imaging data. Our objective was to predict mortality in SAH patients by processing the initial CT scan on a CNN based algorithm.

METHODS: Retrospective multicentric study of a consecutive cohort of patients with SAH between 2011-2022. Demographic, clinical and radiological variables were analyzed. Pre-processed baseline CT scan images were used as the input for training a CNN using AUCMEDI Framework. Our model's architecture leverages the DenseNet-121 structure, employing transfer learning principles. The output variable was mortality in the first three months. Performance of the model was evaluated by statistical parameters conventionally used in studies involving artificial intelligence methods.

RESULTS: Images from 219 patients were processed, 175 for training and validation of the CNN and 44 for its evaluation. 52%(115/219) of patients were female, and the median age was 58(SD=13.06) years. 18.5%(39/219) were idiopathic SAH. Mortality rate was 28.5%(63/219). The model showed good accuracy at predicting mortality in SAH patients exclusively using the images of the initial CT scan (Accuracy=74%, F1=75% and AUC=82%).

CONCLUSION: Modern image processing techniques based on AI and CNN make possible to predict mortality in SAH patients with high accuracy using CT scan images as the only input. These models might be optimized by including more data and patients resulting in better training, development and performance on tasks which are beyond the skills of conventional clinical knowledge.

***Keywords:*** Subarachnoid hemorrhage; Convolutional Neural Networks; Artificial intelligence; Mortality, Prognosis; CT scan

**INTRODUCTION**

Subarachnoid Hemorrhage (SAH) is a devastating form of hemorrhagic stroke with an incidence of 6-8 persons per 100.000 inhabitants per year, with marked variations in specific regions such as Japan, Finland or Indiana[1]. Around 70-80% of spontaneous SAH are caused by the rupture of an intracranial aneurysm, known as aneurysmal SAH (aSAH)[2]. Despite its low incidence, aSAH is a major burden for healthcare systems due to its high mortality and morbidity rates despite optimal treatment[3].

In modern series, 30-day mortality rates range between 27% and 44% and little improvement has been achieved despite extensive efforts to treat its causes or understanding the pathophysiology of the many and treacherous complications that may arise along its course [4]. Predictors of in-hospital mortality include admission clinical grade, rebleeding, delayed cerebral ischemia, treatment-related ischemia, and intraventricular hemorrhage[4,5]. Early brain injury due to the initial hemorrhagic insult and aneurysm rebleeding account for most case fatalities[4]. Therefore, efforts have been addressed to prevent aSAH by controlling vascular risk factors, and to prevent rebleeding by granting an early exclusion of the aneurysm. The latter is the epitome of medical debate, since optimal timing and best therapeutic approach are fiercely discussed[6-8]. However, survival and functional results have scarcely improved during these first decades of the century. Indeed, accurate prediction of outcomes in patients with moderate to poor grades remains a challenge.

Accurate predictions in the medical field often require a large amount of data from large cohorts of patients. Although patients' data is increasingly accessible, managing such complex information has led to the development of modern predictive algorithms and models based on artificial intelligence (AI). Convolutional Neural Networks (CNN), a form of deep learning (DL), mimic the entangled and complex system of connections existing in biological neural structures. Nodes are organized into layers and interconnected to generate and spread output signals resulting from multiple interlinked activation functions. CNN can modify their behavior as they learn from their training. In addition, CNN might consider features or variables otherwise ignored by the observer. CNN have shown excellent performance in accurately predicting various targeted variables in the medical field based on different imaging modalities[9,10]. Some known risk factors for in-hospital mortality associated with aSAH can be identified on the initial CT scan (blood amount, intraventricular hemorrhage, edema, ischemic changes, etc.)[4,5,11,12]. This is advantageous, as a single sequence of images acquired upon admission can provide most of the relevant information necessary to predict a patient's course.

In this clinical investigation we sought to design, create and evaluate a model based on CNN applied to the initial CT scan to predict the mortality at three months of patients admitted to the hospital with a SAH.

**METHODOLOGY**

The present investigation is reported following the Strengthening the Reporting of Observational Studies in Epidemiology(STROBE) guidelines[13] and the CheckList for Artificial Intelligence in Medical imaging (CLAIM)[14]. The study protocol was approved by the Institutional Review Board (22-PI180).

*Study population*

The clinical records of a consecutive cohort of patients diagnosed with SAH admitted to our institution between 2011 and 2022 were retrospectively reviewed. Additionally, a series of consecutive patients from another institution with similar characteristics was included to test the robustness of the algorithm. SAH diagnosis was based on compatible clinical signs and a positive CT scan without previous history of a recent traumatic event. Patients with known survival status at three months suffering both aneurysmal or non-aneurysmal SAH were included. Patients whose CT scans were acquired later than 24h from the onset of symptoms or that could not be properly processed were excluded.

*Variables*

Demographic data such as age and sex; cardiovascular risk factors (Smoking, Hypertension, Diabetes, Dyslipidemia, Familiar History of SAH); admission clinical severity scales [World Federation of Neurosurgical Societies (WFNS) and Hunt and Hess(HH) grading scales]; Modified Fisher(mF) scale and Mortality at 3 months were obtained from the clinical records of included patients[15-17].

*Image Acquisition*

CT scans from the institutional and external cohorts were respectively acquired on a PHILLIPS INGENUITY CT and a SIEMENS SOMATOM CT scanners (Supplementary Table 1).

*Image Preprocessing*

CT images were sourced from the Picture Archiving Communication System (PACS), with the images being in the Digital Imaging and Communications in Medicine (DICOM) format, prior to subsequent manipulation. An initial step involved the transformation of these images into the Neuroimaging Informatics Technology Initiative (NIfTI) format, employing the dicom2niix tool version v1.0.20220720, which can be accessed here:

https://github.com/rordenlab/dcm2niix/releases/tag/v1.0.20220720.

In certain CT scans, the gantry tilt is adjusted to minimize radiation exposure to non-cerebral regions, resulting in image slices being at an inclined angle. To avert complications in 3D operations, such as inaccurate voxel distances between slices, the corrected image output from dcm2niix was utilized for further manipulations.

Although Hounsfield Units (HU) are the conventional format for CT analysis, negative HU values could potentially pose problems with standard imaging pipelines developed for Magnetic Resonance Imaging (MRI), which generally exhibit positive values. To mitigate this, we implemented an intensity normalization through a lossless transformation to Cormack units, which have a minimum value set at zero. This step was performed using the Clinical Toolbox for SPM available at https://github.com/neurolabusc/Clinical .

Subsequently, we conducted a brain extraction procedure using the Brain Extraction Tool (BET) from the Functional MRI of the Brain Software Library (FSL) v6.0, which can be found here: https://fsl.fmrib.ox.ac.uk/fsl/fslwiki/BET/UserGuideThe final step involved the registration to an CT template image of dimensions 1 mm x 1 mm x 1 mm, and 193, 229, 193 size. This was achieved through the application of diffeomorphic registrations using symmetric normalization (SyN) from Advanced Normalization Tools (ANTs), available at https://github.com/ANTsX/ANTs.

*Neural Network*

In our research, the AUCMEDI-Framework (https://frankkramer-lab.github.io/aucmedi/) [18] was used to instruct a deep neural network in differentiating between two patient outcomes: survival and death.

*Architecture*

A DenseNet121, a derivative of the Dense Convolutional Network (DenseNet) was implemented [19]. This DL architecture stands out due to its dense connectivity pattern, its computational efficiency and minimal memory usage, which stems from the reutilization of features. It was selected for its proficiency in extracting intricate and hierarchical features from input images, a critical component in medical image analysis (Figure 1).

*Activation Output*

For our binary classification task, the *softmax* function was used as the final activation function. It generates two probabilities that sum up to one, representing the likelihood of the input image belonging to each of the two classes. The class with the higher probability is chosen as the output prediction.

*Class Imbalance and Loss Function*

We initially calculated the class weights to be implemented by the categorical focal loss function. This function prioritizes harder-to-classify instances and downplays simpler examples, thereby guiding the model to concentrate more on a balanced set of challenging samples. This method enhances the model's performance on imbalanced datasets.

*Data Augmentation*

In our CNN model, we employed several image augmentation techniques to increase the diversity and robustness of our training dataset. These techniques included mirroring (reflecting images across their vertical or horizontal axis), rotation (adjusting images by a certain degree around the center point), scaling (changing the size of the images), and elastic transformation (distorting the image locally by displacing each pixel randomly, which simulates natural variations). These augmentations increase the potential generalization of the model as they allow for a better performance on unseen data.

*Callbacks*

In our model, we employed several Callbacks including *EarlyStopping, ModelCheckpoint*, and *ReduceLROnPlateau*. *EarlyStopping* is used to halt training when a monitored metric has stopped improving, preventing overfitting and saving computational resources. *ModelCheckpoint* allows for the saving of the model after each epoch, ensuring the retention of the best performing model. *ReduceLROnPlateau,* on the other hand, lowers the learning rate when a metric has ceased to improve, optimizing the model's ability to find the global minimum and enhance the training performance. These strategies work together to mitigate overfitting and reduce unnecessary training time.

*Transfer Learning*

Transfer learning is a machine learning(ML) approach that applies a pre-trained model, typically developed on extensive datasets, to a new but related task. This strategy bolsters learning efficiency, especially when data for the new task is limited. The model retains or 'freezes' the learned weights from the prior task while fine-tuning the classification layer to the new task. After several epochs, the model is fully unfrozen for additional fine-tuning, thereby conserving computational resources and training time.

*Explainable Artificial Intelligence*

We employed Gradient-weighted Class Activation Mapping (Grad-CAM) for explainable artificial intelligence[20]. This technique provides visual elucidations for decisions made by CNNs. It uses the gradients of any targeted concept, flowing into the final convolutional layer, to generate a coarse localization map that emphasizes the crucial regions in the image for predicting the result. This method enhances model´s interpretability and transparency, enabling users to comprehend the critical areas in the input that primarily influence the model's decision.

**Metadata**

A second CNN predictive model was developed, incorporating baseline CT scan images and admission-related clinical information as input. The clinical data was limited to variables available upon admission, such as Age, Sex, Hypertension, WFNS grade, Acute Hydrocephalus, and others. The purpose of creating this model was to determine whether baseline CT scan images alone could provide a robust predictive model or if the addition of clinical information could enhance its performance. Only variables demonstrating a statistically significant association with mortality were included.

**Statistics**

Excel (Microsoft. Redmon,WA.USA.16.16.4version) and SPSS Statistics (IBM.Armonk, NY.USA 24 version) were implemented to run conventional statistical methods. The distribution of continuous variables was assessed using a normality test. Categorical variables are expressed in frequencies and percentages. Categorical variables were compared with Chi square and exact Fisher tests. Association between mortality and continuous variables was analyzed with Student´s T-Test or Wilcoxon U-test. Univariate analysis was performed to study the association of clinical variables with mortality at three months. Performance of CNN was evaluated by metrics typically implemented in DL methods: Sensibility, Specificity, Accuracy, F1 score and Area Under the Curve (AUC) for Receiver Operating Characteristic (ROC) curve.

**RESULTS**

A total of 219 patients met the inclusion criteria for the study (Figure 2). Among them, 47.5% (104/219) were males, and the mean age was 58 (SD= 13.06). A perimesencephalic pattern on the initial CT scan was observed in 37 patients (16.9%). In 42 cases, the initial arteriography did not detect the presence of an aneurysm, out of which 39 cases (17.8%) were confirmed as idiopathic SAH.

Aneurismatic SAH accounted for 180 patients with 222 aneurysms and 36 cases (20%) of multiple aneurysms. The mean WFNS and HH on admission were respectively 2.5 (SD=1.6) and 2.2 (SD=1.6), with a mode, in both cases of 2. Mean mF scale was 3.3(SD=0.9) and the mode was 4. In aSAH, 91 (50.5%) patients were treated surgically, 72 (40%) endovascularly and 17 (9%) were not treated due to brain death signs prior to being possible to provide any effective treatment. In 54.6 % of treated patients, the aneurysm was excluded in the first 24h after the diagnosis. Rebleeding occurred in 15 patients and only 4 of them survived. In the sample of 219 patients, mean stay was 24 days and mortality rate was 28.5 % (Table 1).

Among patients with SAH, mortality was significantly superior in older individuals (61.2 vs 56.5 years old, F=5.12, t=2.48; p=0.014), female patients (35.6% vs 23.1%; $X^2$=4.14; p=0.042), and patients with hypertension (40.6% vs 21.1% , $X^2$=9.81; p=0.002), intraparenchymal hematoma (48.4% vs 21.9%, $X^2$=15.24; p<0.001) and acute hydrocephalus (42.7% vs 19.8%, $X^2$=13.04; p<0.001). Patients with higher grades in modified Fisher ($X^2$=39.9; p<0.001), WFNS ($X^2$=46.9; p<0.001) and HH ($X^2$=48.6; p<0.001) scales experienced higher mortality rates (Figure 1). All these variables were included as metadata in the CNN model based on baseline CT scan images and clinical information. Other cardiovascular risk factors like Diabetes, Dyslipidemia or Smoking were not associated with a higher risk of mortality. Subdural Hematoma or seizures on admission were neither associated with mortality.

Highest grades on WFNS, HH and mF scales demonstrated a strong association with mortality. Remarkably, mF grades 3 or 4 proved to be a strong risk factor for mortality compared to mF grade 1 or 2: Odds Ratio of 21.7 (p=0.003; 95% Confidence Interval: 2.91 – 161.71). The results for other variables are shown on Table 2.

CNN algorithms were developed, trained, validated, and tested in this study. Among the models created, the one exclusively based on the initial CT scan demonstrated the best performance. The optimal performance was achieved during the final epoch, with the following metrics: Sensitivity = 0.75(SD=0.025; CI95%= 0.716-0.786), Specificity = 0.75 (SD=0.025; CI95%= 0.716-0.786), Accuracy = 0.74(SD=0), F1 score = 0.72(SD=0.025; CI95%= 0.615-0.829, and AUC (Area Under the Curve) = 0.82(SD=0). The inclusion of additional clinical metadata to the model did not significantly enhance its performance. The best F1 score obtained with the combined model was as follows: Sensitivity = 0.75(SD=0.025; CI95%= 0.716-0.786), Specificity = 0.75(SD=0.025; CI95%= 0.716-0.786), Accuracy = 0.74(SD=0), F1 score = 0.74(SD=0.077; CI95%= 0.663-0.817), and AUC = 0.80(SD=0). Results are presented in Table 3 and depicted on figures 2-4. (Supplementary Table 2).

**DISCUSION**

In this investigation we retrospectively reviewed all consecutive cases of SAH CTs patients admitted to our institution and we validated our results with an external cohort from another center. Images and data were preprocessed and used to train a CNN to predict mortality in a test cohort of patients. Results demonstrated that a CNN predictive algorithm exclusively based on the initial CT outperforms the combination of images and clinical data. The results of this image-based algorithm proved the ability of CNN to establish solid predictions using medical images as input. To the best of our knowledge, this study represents the first successful development of an image-based CNN algorithm that accurately predicts mortality in patients with SAH.

Several studies have demonstrated the ability of DL models to identify abnormalities on head CT scans: Hemorrhage, fractures, stroke, edema, etc [21,22]. These investigations require extensive labelled image datasets as inputs to build the model[21]. Different approaches have been used for this purpose, from classifying slices as pathologic or normal, to automatic segmentation of abnormal areas[22-25]. Using AI models to accomplish iterative and tedious tasks such as blood segmentation is a significative advancement that reduces working times and allows to process large samples of patients increasing the statistical power of clinical investigation. However, in most of the available scientific papers regarding SAH or brain hemorrhage, automated processes are limited to feature extraction. Those features are then used in conventional statistical methods or ML algorithms but a fully automated pipeline capable of accurately predict a clinical outcome from raw images was lacking for SAH. In this sense, our DL model represents a further leap forward.

Regarding SAH, efforts have also focused on aneurysm detection. Using different modalities of images (CT, DSA or MRI) and approaches (Stand-alone AI or AI supporting a clinician) several reports have demonstrated the ability of AI to assist in aneurysm detection[26]. Bo et al. demonstrated the utility of a DL-based model to assist radiologists in the detection of intracranial aneurysms using AngioCT [27]. Increasing the reliability and, particularly, the specificity of these automated models could allow for future screening of intracranial aneurysms in large populations in a context in which human intervention could be relegated to supervision and final confirmation of positive results.

Mortality and outcome prediction has classically relied on risk factors and clinical and radiological scales. Advanced methods of data processing represent a great opportunity to exploit the information patients harbor early on admission. Therefore, studies have implemented ML methods to extract the best from features with known implication on the final outcome. Dengler et al. compared the performance on outcome prediction for aSAH patients of ML methods and established clinico-radiological scores[28]. Authors found that GCS and Age were the most relevant features for outcome prediction and that ML methods were not superior to conventional scores[28]. In a study based on clinical features and ML methods, Toledo et al. achieved an AUC for the ROC curve of 0.85 in a decision tree built with Fisher and WFNS scales to predict functional outcome [29]. Later on, Lo et al. used an extensive database to create an outcome predictive algorithm[30]. This model was based on multiple demographic and clinical variables that were used as the input for Bayesian CNN with Fuzzy Logic Inferences[30]. The AUC for the ROC curve was 0.85. However, many of the features that fed the algorithm were not present on admission, therefore an early prediction of patient´s outcome was not feasible[30]. Our model has the ability to make predictions without any clinical input or need for expert assessment, which has the full potential for automatization, generalization and applicability in primary and secondary centers referring patients to tertiary hospitals.

A paradoxical finding of our research is the null improvement of the predictive model with the addition of clinical metadata with otherwise proven association with mortality. It was hypothesized that CNN would extract information from the images beyond human capacity, but we also expected that clinical data would have improved the model. Probably, the image-based model is able to estimate the quantity and distribution of blood, detect signs of brain damage such as edema, herniation and effacement of basal cisterns. Many of these radiological signs are known factors of bad clinical grade on admission and have previously been correlated with mortality[4,11,12]. Previous works in other areas have highlighted how the clinical information adds up to an image-based model, while other groups demonstrated exactly the opposite[31-33]. These conflicting experiences might be due to differences on the targeted prediction, the architecture of the NN or, even, on how the clinical information is introduced into the model. Nonetheless, it is also possible that the baseline CT scan harbors highly valuable information that conventionally considered regarding final outcome. This fact would challenge the idea of the influence of clinical management and delayed cerebral lesion on SAH mortality and emphasize the relevance of initial damage in final outcome.

**Limitations**

The present study harbors some limitations. First, predictions are based on a ground truth which, in this case, derives from the results of our practice. Mortality rates, causes of death, risk factors, treatment choices and overall management may vary amongst institutions. This flaw can only be tackled if larger training cohorts from different institutions, representing different managing protocols are used to build the model. Efforts should be made in this regard if we aim for a predictive tool that can be applied to as many healthcare contexts as possible. Second, CT scanners protocols and manufacturers might change the information the model extracts from the image and consequently the class assigned to a particular case. However, the methodology we implemented to harmonize CT scans was designed to minimize variability in the imaging data, improve the robustness of the training and the accuracy and reliability of our model. Third, it can be argued that perimesencephalic SAH and aSAH are completely different diseases, with vast differences on clinical evolution and final outcome, and therefore, should not be mixed in a mortality prediction study. Trained specialists in neurovascular emergencies might correctly identify a SAH as perimesencephalic with a rapid view of the CT scan and short assessment of the patient. However, one of the main applications of an image-based model as the one herein presented is to support clinicians on their decisions, especially in non-tertiary centers where knowledge about alarm signs and prognosis might be scarce. Finally, one of the main problems of DL is that the algorithm does not disclose what their decisions were based on, in other words, we cannot fully explain why the model classifies a given case into a specific class. Efforts are being made to unlock the black box that DL methods often represent. These efforts are referred as Explainable Artificial Intelligence or XAI. In DL, XAI methods are mainly post-hoc, meaning that the trained model is analyzed to find learned associations[34]. Our team is currently working on visual activation maps or saliency maps based on Gradient-weighted class activation mapping *(Grad-CAM)*, which are graphic representations of the areas of an image that are important for the model to make a decision or classify a case into a group[20,35](Figure 5-6).

**CONCLUSION**

DL algorithms based on initial CT scan allow to provide accurate predictions on mortality for SAH patients. The limited improvement seen with the addition of clinical information suggests that many factors influencing patient outcome are present in the early stages of the disease and could be identified on the initial CT scan. AI predictive models are a promising tool that could significantly improve the understanding and decision-making process in complex pathologies like SAH. However, further optimization of these models through inclusion of more data and patients is necessary to enhance their performance on complex tasks which are beyond the potential of conventional clinical knowledge.

**Code availability**

The source code for the AUCMEDI framework can be found at https://github.com/frankkramer-lab/aucmedi

Additionally, the pipeline utilized in our study will be made publicly available at https://github.com/smcch/HSA_CNN

These repositories provide access to the respective source codes, enabling researchers and interested individuals to explore and utilize the frameworks and pipelines implemented in our study.

**TABLES**

| VARIABLE | MEAN/MODE | NUMBER | PERCENTAGE |
|---|---|---|---|
| **Female** | | 115 | 52.5% |
| **Age** | 57.9 (SD=13.06)years | | |
| **RISK FACTORS** | | | |
| **HT** | | 96 | 43.8 % |
| **Tobacco** | | 90 | 41% |
| • Smoker Female | | 40 | 34.8% of females |
| • HT + Tobacco | | 36 | 16.5 % of total |
| **Diabetes** | | 20 | 9% |
| **Dislypidemia** | | 81 | 37% |
| **Familial History** | | 3 | 1.5% |
| **Idiopathic SAH** | | 39 | 18% |
| **Aneurysmal SAH** | | 180 | 82% |
| **Multiple** | | 36 | 20% |
| **Anterior Circulation** | | 174 | 97% |
| **Aneurysm Diameter** | 7.9 mm (SD=5.6) | | |
| **TREATMENT** | | | |
| **Surgical** | | 91 | 50.5% |
| **Endovascular** | | 72 | 40% |
| **No treatment** | | 17 | 9% |
| **Timing of treated** | | | |
| • Ultra Early(<24h) | | 114 | 70% |
| • Early (24-72h) | | 33 | 20% |
| • Delayed(>72h) | | 16 | 10% |
| **ADMISSION** | | | |
| **Hunt & Hess** | 2.2/2 | | |
| I | | 103 | 47% |
| II | | 45 | 20.5% |
| III | | 10 | 4.5% |
| IV | | 17 | 8% |
| V | | 44 | 20% |
| **WFNS** | 2.5/2 | | |
| I | | 93 | 42.5% |
| II | | 46 | 21% |
| III | | 8 | 3.5% |
| IV | | 26 | 12% |
| V | | 46 | 21% |
| **Modified Fisher** | 3.2/4 | | |
| I | | 15 | 7% |
| II | | 25 | 11.5% |
| III | | 35 | 16% |
| IV | | 144 | 65.5% |
| **Intraparenchymal hematoma** | | 63 | 35% |
| **Subdural Hematoma** | | 9 | 5% |
| **COMPLICATIONS** | | | |
| **Acute Hydrocephalus** | | 96 | 44% |
| **Shunt Dependent Hydrocephalus** | | 38 | 17.5% |
| **Seizure** | | 37 | 17% |
| **Epilepsy** | | 12 | 6.5% |
| **Symptomatic Vasospasm** | | 38 | 17.5% |
| **Delayed Cerebral Ischemia** | | 52 | 23.5% |
| **Length of Stay** | 24 days | | |
| **OUTCOME** | | | |
| mRS at 3 months | 3/6 | | |
| 0 | | 46 | 21% |
| 1 | | 37 | 17% |
| 2 | | 16 | 7.5% |
| 3 | | 19 | 8.5% |
| **4** | | 15 | 7% |
| **5** | | 23 | 10.5% |
| **6** | | 63 | 28.5% |
| **Mortality** | | 63 | 28.5% |

**Table 1.** Characterization of the sample according to analyzed variables.

HT: Hypertension; mRS: Modified Rankin Scale; WFNS: World Federation of Neurosurgeons.

| Variable | Reference Value | Degrees of Freedom | p | OR | 95% CI |
|---|---|---|---|---|---|
| Sex (Male) | Male | 1 | 0.042 | 0.54 | 0.30-0.98 |
| Age | 1 | | 0.014 | 1.03 | 1.01-1.05 |
| Hypertension | Yes | 1 | 0.002 | 2.55 | 1.41-4.63 |
| Intraparenchimatous Hematoma | Yes | 1 | <0.001 | 3.34 | 1.79-6.22 |
| Acute Hydrocephalus | Yes | 1 | <0.001 | 3.01 | 1.64-5.55 |
| WFNS | 1 | 4 | <0.001 | | |
| WFNS 2 | | | 0.007 | 3.67 | 1.44-9.42 |
| WFNS 3 | | | 0.033 | 5.60 | 1.14-27.40 |
| WFNS 4 | | | 0.001 | 5.83 | 2.05-16.62 |
| WFNS 5 | | | <0.001 | 17.50 | 6.99-43.77 |
| Hunt & Hess | 1 | 4 | <0.001 | | |
| HH 2 | | | 0.002 | 4.51 | 1.72-11.86 |
| HH 3 | | | 0.047 | 5.00 | 1.02-24.51 |
| HH 4 | | | 0.02 | 4.63 | 1.27-16.87 |
| HH 5 | | | <0.001 | 25.00 | 8.99-69.56 |
| Modified Fisher. Dichotomized | <2 | 1 | <0.001 | | |
| mF >2 | | | 0.003 | 21.70 | 2.91-161.72 |

**Table 2.** Odds Ratio for variables demonstrating statistically significant association with mortality.

| MODEL | IMAGE-BASED NEURAL NETWORK PERFORMANCE | | | | IMAGE AND METADATA BASED NEURAL NETWORK PERFORMANCE | | | |
|---|---|---|---|---|---|---|---|---|
| EPOCH METRIC | BEST AUC | BEST F1 | BEST LOSS | LAST | BEST AUC | BEST F1 | BEST LOSS | LAST |
| TP | 14.5 | 15 | 15 | **16** | 16 | 16.5 | 15.5 | 15 |
| TN | 14.5 | 15 | 15 | **16** | 16 | 16.5 | 15.5 | 15 |
| FP | 7 | 6.5 | 6.5 | **5.5** | 5.5 | 5 | 6 | 6.5 |
| FN | 7 | 6.5 | 6.5 | **5.5** | 5.5 | 5 | 6 | 6.5 |
| Sensitivity | 0.53 | 0.61 | 0.74 | **0.75** | 0.60 | 0.75 | 0.69 | 0.50 |
| Specificity | 0.53 | 0.61 | 0.74 | **0.75** | 0.60 | 0.75 | 0.69 | 0.50 |
| Precision | 0.56 | 0.63 | 0.70 | **0.72** | 0.75 | 0.73 | 0.68 | 0.35 |
| FP Rate | 0.47 | 0.39 | 0.26 | **0.25** | 0.40 | 0.25 | 0.31 | 0.50 |
| FN Rate | 0.47 | 0.39 | 0.26 | **0.25** | 0.40 | 0.25 | 0.31 | 0.50 |
| FDR | 0.44 | 0.37 | 0.30 | **0.28** | 0.25 | 0.27 | 0.32 | 0.15 |
| Accuracy | 0.67 | 0.70 | 0.70 | **0.74** | 0.74 | 0.77 | 0.72 | 0.70 |
| F1 | 0.51 | 0.61 | 0.69 | **0.72** | 0.60 | 0.74 | 0.68 | 0.41 |
| AUC | 0.72 | 0.74 | 0.73 | **0.82** | 0.78 | 0.80 | 0.78 | 0.35 |

**Table 3**. Performance of Neural Networks Algorithms. Results are presented as average of both analyzed classes (Dead or alive at three months follow up).

FDR: False Discovery Rate; FN: False Negative; FP: False Positive; TN: True Negative; TP: True Positive

**FIGURES**

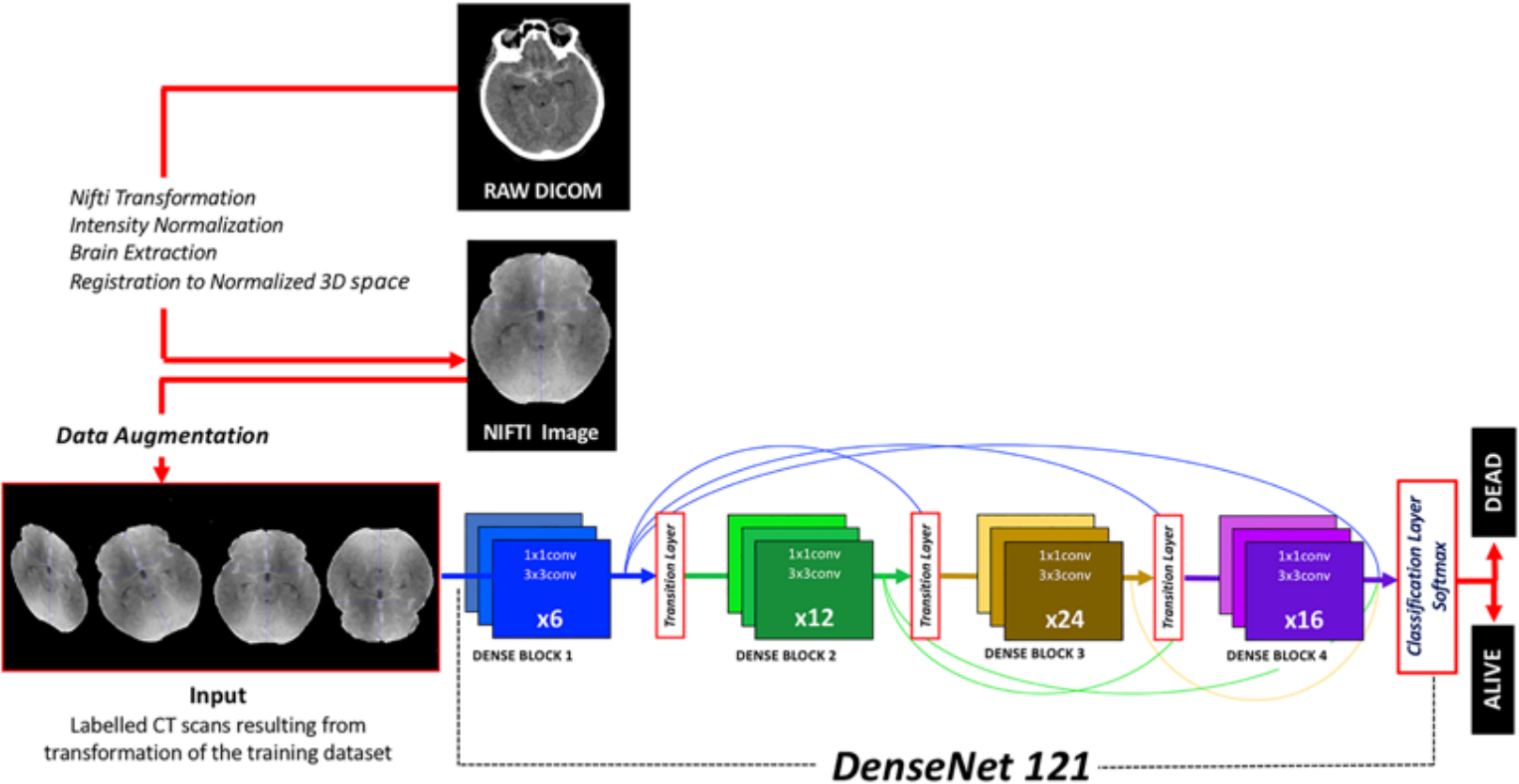

**Figure 1**. Diagram depicting the workflow from DICOM raw images to classification results provided by the algorithm. Image preprocessing, data augmentation, neural network architecture and output classification function are herein represented.

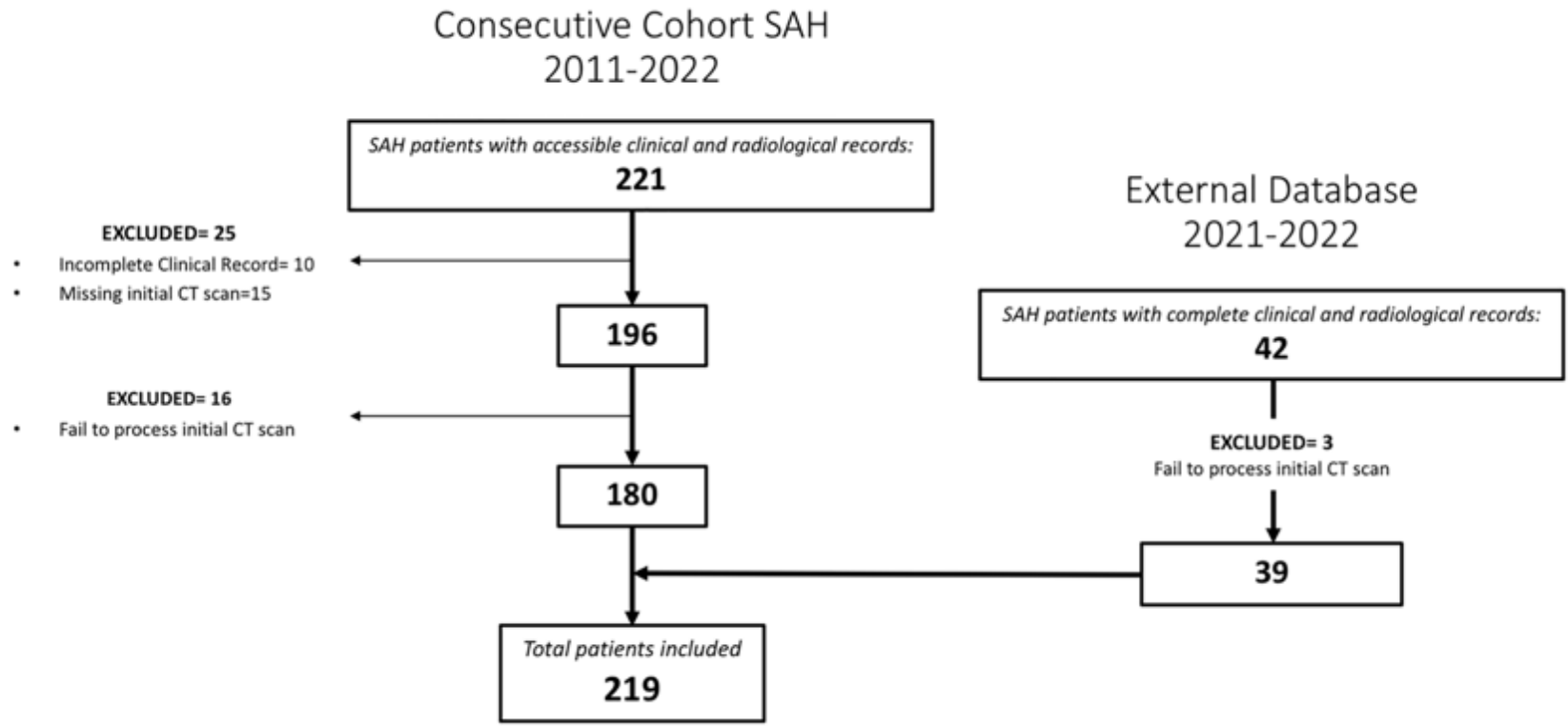

**Figure 2**. Flowchart describing the screened, included and excluded patients for the institutional and external cohort of patients whose image and data were implemented to create, train and evaluate the model.

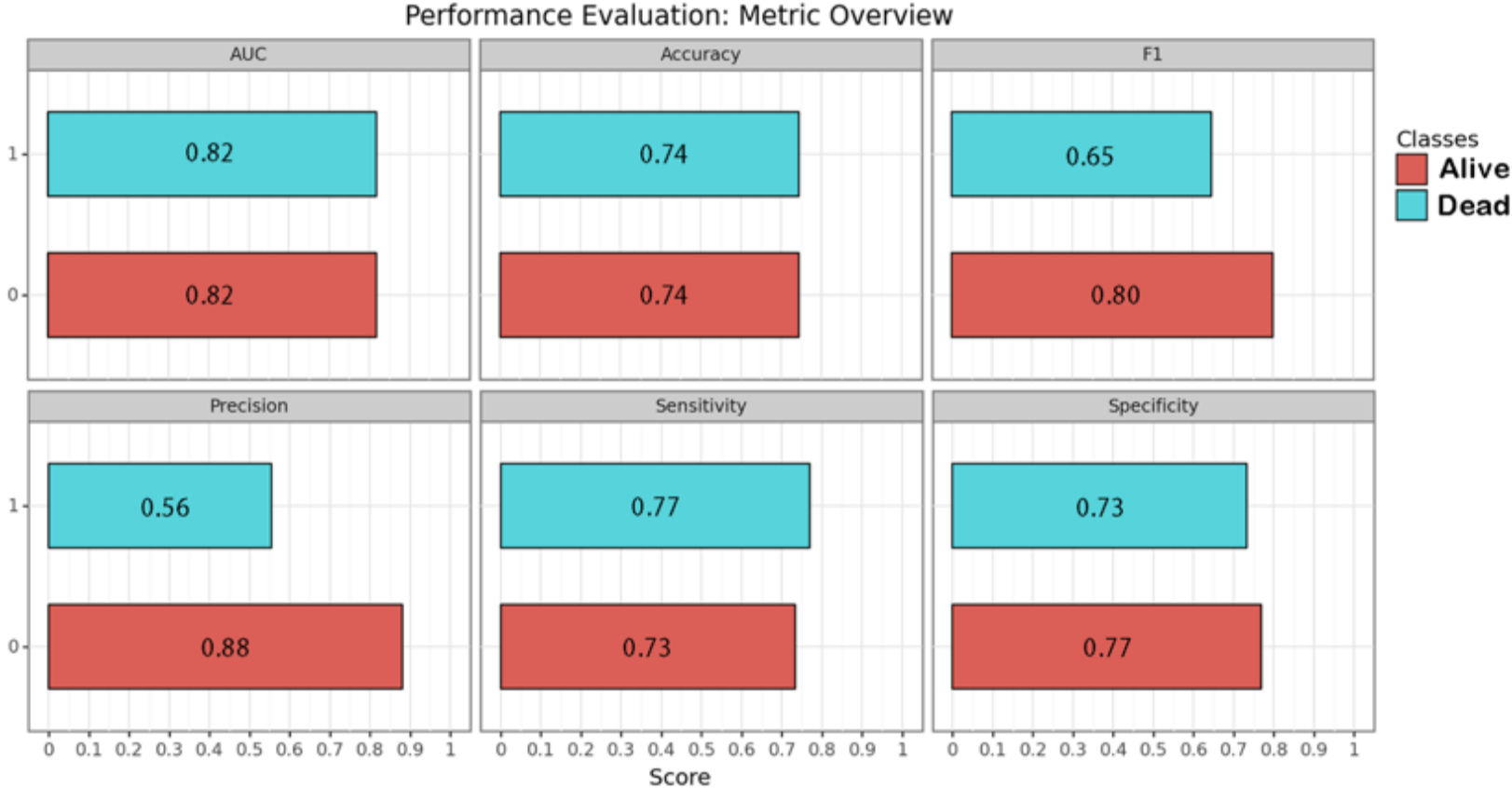

**Figure 3.** Performance of the Image-based Neural Network Algorithm. Each metric is represented for each class by its correspondent bar and the numeric value within.

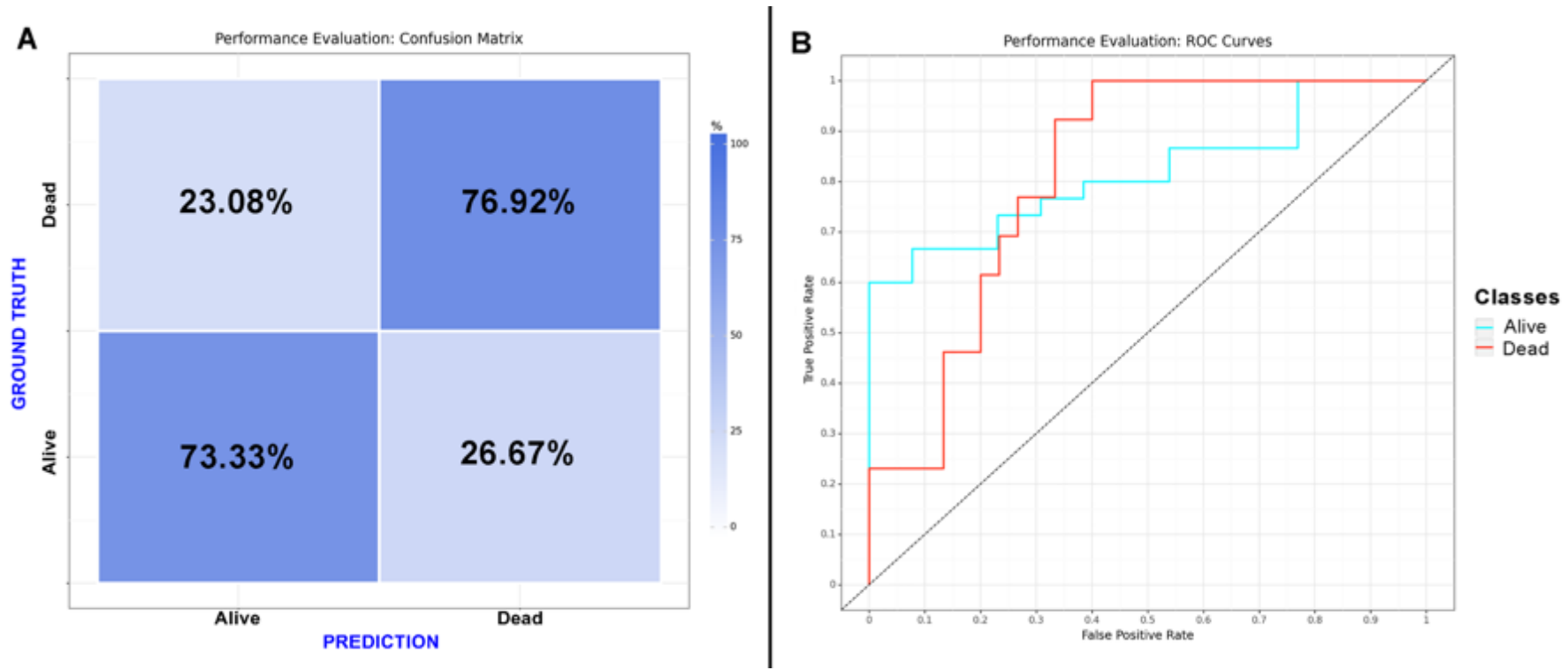

**Figure 4.** Performance Evaluation. **A.** Confusion matrix of the CNN, considering ¨dead¨ as a positive result to the test. **B.** Receiver Operating Characteristic Curve of the CNN.

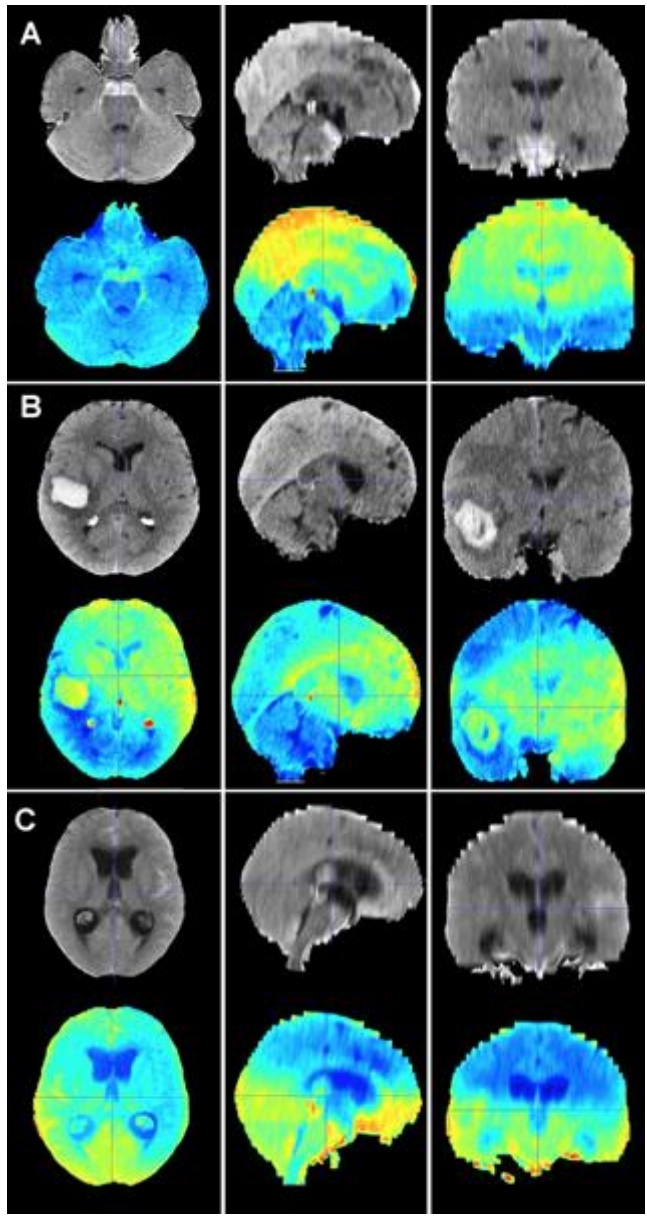

**Figure 5.** Baseline CT scan (upper row) and Gradient class activation maps or Grad-CAM(lower row) for three patients (A,B,C) on the test cohort who were alive three months after suffering a subarachnoid hemorrhage. Saliency maps highlight regions in red that are more significant to classify patients, in this case, into the group of patients who survived the event. Thus, it is possible to create a visual depiction of the process the model follows to allocate patients into each class. These maps highlight supratentorial brain areas and seem to disregard hemorrhage except when it follows a perimesencephalic pattern. **A**. 60 years old male who suffered a perimesencephalic SAH whose Angio-MR and Angiography were negative. **B.** 43 years old male who was diagnosed with a SAH caused by a right Middle Cerebral Artery artery aneurysm. Admitted in good condition (WFNS 1) was surgically treated and discharged without major neurological deficits on postoperative day 19. **C**. 75 years old female who suffered a SAH and was admitted to the hospital on WFNS grade 4. The left posteroinferior cerebellar artery aneurysm was coiled. The patient survived the event but was still severely impaired at three months follow up (Modified Rankin Scale: 4).

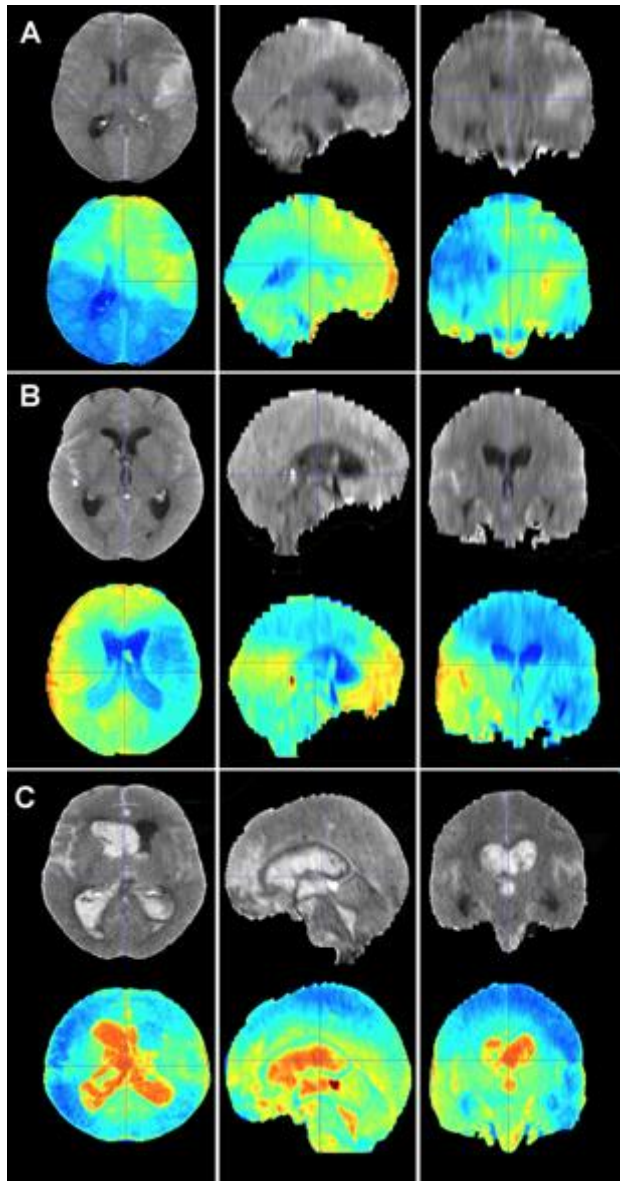

**Figure 6.** Baseline CT scan (upper row) and Gradient class activation maps or Grad-CAM(lower row) for three patients (A,B,C) on the test cohort who died as a result of a subarachnoid hemorrhage(SAH). These maps visually illustrate the areas the model considers to allocate patients into the "dead" group. Grad-CAM maps show that posterior fossa and intraventricular and cisternal blood might be relevant areas or items to consider in order to classify patients as dead. **A**. 51 years old male who suffered a SAH due to the rupture of a left middle cerebral artery aneurysm. The patient who initially presented with a WFNS grade 2, abruptly deteriorated to a WFNS grade 5 requiring emergent surgical treatment. The patient died on postoperative day 56 as a consequence to both systemic and neurological complications. **B.** 70 years old female who was diagnosed with a SAH caused by a right Posterior Communicating artery aneurysm who died 40 days after her admission due to a combination of factors including delayed cerebral ischemia, meningitis and pneumonia. **C.** 78 years old male with a SAH caused by an Anterior communicating artery admitted to the hospital with a WFNS grade 5 and a mF grade 4 who died the next day to the event.

SUPLEMENTARY MATERIAL

SUPPLEMENTARY MATERIAL

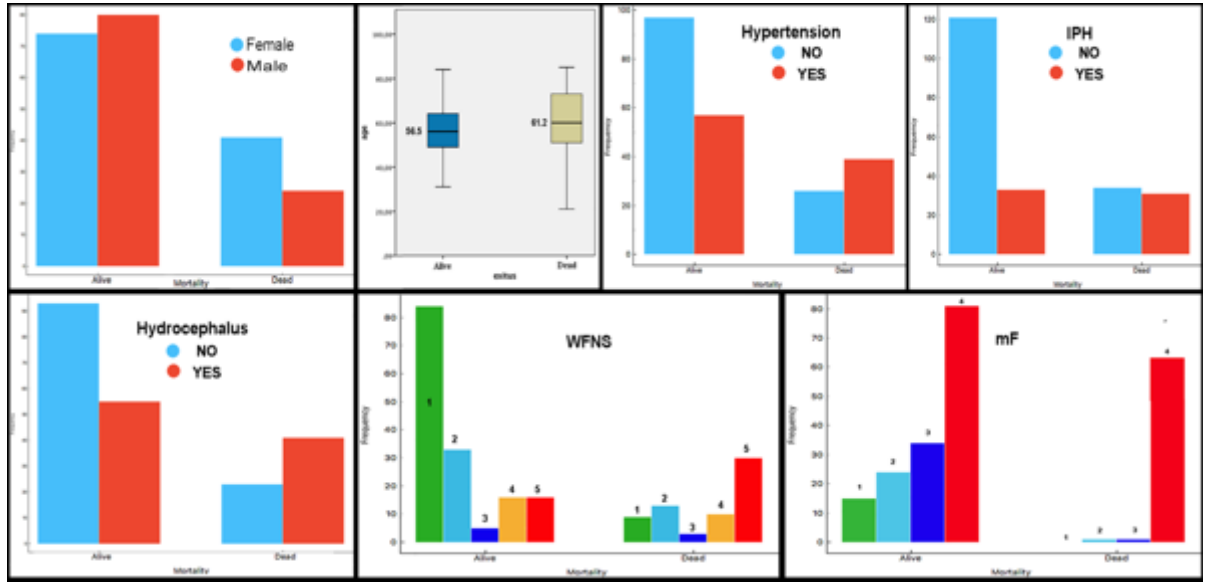

**Supplementary Figure 1:** Distribution of mortality according to Risk factors: Sex, Age, Hypertension (HT), Intraparenchymal hematoma (IPH), WFNS and Modified Fisher (mF).

| Manufacturer | PHILIPS | SIEMENS |
|---|---|---|
| Model | Ingenuity CT | Somatom |
| Scan Mode | Spiral | Spiral |
| Tube Voltage (kV) | 120 | 120 |
| Exposure(mAs) | 189-221 | 199-262 |
| Average CTDIvol (mGy) | 48 | 57 |
| Rotation Time (s) | 0.4 | 1 |
| Collimation | 64x 0.625 | 64x0.625 |
| Pitch factor | 0.3 | 0.5 |
| Slice Thickness(mm) | 3-4 | 2-4 |
| Number of slices | 64-99 | 66-150 |
| Field of view (mm) | 250 | 250 |
| Samples per pixel | 1 | 1 |
| CT detector array | 512x512 | 512x512 |

**Supplementary Table 1**. CT scan protocol parameters.

*CTDIvol: Computed Tomography Dose Index volume.*

| CLASS | METRIC | IMAGE-BASED NEURAL NETWORK PERFORMANCE | | | | IMAGE AND METADATA BASED NEURAL NETWORK PERFORMANCE | | | |
|---|---|---|---|---|---|---|---|---|---|
| | | BEST AUC | BEST F1 | BEST LOSS | LAST | BEST AUC | BEST F1 | BEST LOSS | LAST |
| ALIVE | TP | 27 | 25 | 19 | 22 | 29 | 24 | 23 | 30 |
| | TN | 2 | 5 | 11 | 10 | 3 | 9 | 8 | 0 |
| | FP | 11 | 8 | 2 | 3 | 10 | 4 | 5 | 13 |
| | FN | 3 | 5 | 11 | 8 | 1 | 6 | 7 | 0 |
| | Sensitivity | 0.90 | 0.83 | 0.63 | 0.73 | 0.97 | 0.80 | 0.77 | 1 |
| | Specificity | 0.15 | 0.38 | 0.85 | 0.77 | 0.23 | 0.69 | 0.62 | 0 |
| | Precision | 0.71 | 0.76 | 0.90 | 0.88 | 0.74 | 0.86 | 0.82 | 0.70 |
| | FPR | 0.85 | 0.62 | 0.15 | 0.23 | 0.77 | 0.31 | 0.38 | 1 |
| | FNR | 0.10 | 0.17 | 0.37 | 0.27 | 0.03 | 0.20 | 0.23 | 0.00 |
| | FDR | 0.29 | 0.24 | 0.10 | 0.12 | 0.26 | 0.14 | 0.18 | 0.30 |
| | Accuracy | 0.67 | 0.70 | 0.70 | 0.74 | 0.74 | 0.77 | 0.72 | 0.70 |
| | F1 | 0.79 | 0.79 | 0.75 | 0.80 | 0.84 | 0.83 | 0.79 | 0.82 |
| | AUC | 0.72 | 0.74 | 0.73 | 0.82 | 0.78 | 0.80 | 0.78 | 0.35 |
| DEAD | TP | 2 | 5 | 11 | 10 | 3 | 9 | 8 | 0 |
| | TN | 27 | 25 | 19 | 22 | 29 | 24 | 23 | 30 |
| | FP | 3 | 5 | 11 | 8 | 1 | 6 | 7 | 0 |
| | FN | 11 | 8 | 2 | 3 | 10 | 4 | 5 | 13 |
| | Sensitivity | 0.15 | 0.38 | 0.85 | 0.77 | 0.23 | 0.69 | 0.62 | 0 |

| | Specificity | 0.90 | 0.83 | 0.63 | 0.73 | 0.97 | 0.80 | 0.77 | 1.00 |
|---|---|---|---|---|---|---|---|---|---|
| | Precision | 0.40 | 0.50 | 0.50 | 0.56 | 0.75 | 0.60 | 0.53 | - |
| | FPR | 0.10 | 0.17 | 0.37 | 0.27 | 0.03 | 0.20 | 0.23 | 0 |
| | FNR | 0.85 | 0.62 | 0.15 | 0.23 | 0.77 | 0.31 | 0.38 | 1 |
| | FDR | 0.60 | 0.50 | 0.50 | 0.44 | 0.25 | 0.40 | 0.47 | - |
| | Accuracy | 0.67 | 0.70 | 0.70 | 0.74 | 0.74 | 0.77 | 0.72 | 0.70 |
| | F1 | 0.22 | 0.43 | 0.63 | 0.65 | 0.35 | 0.64 | 0.57 | 0 |
| | AUC | 0.72 | 0.74 | 0.73 | 0.82 | 0.78 | 0.80 | 0.78 | 0.35 |

**Supplementary Table 2**. Performance of Neural Networks Algorithms. Results for classes.

FDR: False Discovery Rate; FN: False Negative; FP: False Positive; R: Rate; TN: True Negative; TP: True Positive